\documentclass{article}
\usepackage{ijcai25}
\usepackage{xcolor}

\usepackage{times}
\usepackage{soul}
\usepackage{url}
\usepackage[hidelinks]{hyperref}
\usepackage[utf8]{inputenc}
\usepackage[small]{caption}
\usepackage{graphicx}
\usepackage{amsmath}
\usepackage{amsthm}
\usepackage{booktabs}
\usepackage{algorithm}
\usepackage{algorithmic}
\usepackage[switch]{lineno}
\usepackage{arydshln} 
\usepackage{siunitx} 
\usepackage{amsmath}
\usepackage{amssymb}
\usepackage{multirow}

\title{A Time Series Multitask Framework Integrating a Large Language Model, Pre-Trained Time Series Model, and Knowledge Graph}

\author{ Shule Hao$^1$\and
Junpeng Bao$^{1\ast}$\and
Chuncheng Lu$^{1}$
    \affiliations
    $^1$Xi’an Jiaotong University, Xi’an, China
    \emails
    haoshule@stu.xjtu.edu.cn, baojp@mail.xjtu.edu.cn,  luchuncheng9@stu.xjtu.edu.cn
}

\begin{document}

\maketitle

\begin{abstract}
    Time series analysis is crucial in fields like finance, transportation, and industry. However, traditional models often focus solely on temporal features, limiting their ability to capture underlying information. This paper proposes a novel time series multitask framework, called LTM, which integrates temporal features with textual descriptions to enhance analytical and predictive capabilities. LTM combines pre-trained time series model, large language model (LLM), and knowledge graph to tackle time series tasks, including forecasting, imputation, and anomaly detection. LTM achieves improved performance with a few trainable parameters. It is very efficient and practical.

LTM encodes time series data into patches and enriches user-provided prompts using knowledge graphs to generate enhanced prompts. A novel feature fusion method embeds prompts into each patch encoding, which is processed by a frozen LLM, followed by a feature enhancement module and a time decoder module. During fine-tuning stage, cosine similarity between prompts and temporal patches is integrated into the loss function to boost performance. Experiments on benchmark datasets show that LTM significantly outperforms existing methods. It provides a robust and versatile solution for time series tasks.

\end{abstract}

\section{Introduction}

Time series tasks, including forecasting, imputation, and anomaly detection\cite{liutimer}, are widely applied in domains such as traffic management, energy optimization, and financial forecasting. Traditional time series models primarily focus on mining data patterns but often neglect the semantic context and background information of the task. Their reliance on domain-specific knowledge and task-specific models limits their generality and efficiency.

In contrast, large language models (LLMs), such as ChatGPT \cite{brown2020language}, ChatGLM \cite{glm2024chatglm}, and LLaMA \cite{touvron2023llama}, have demonstrated remarkable generalization capabilities in natural language processing tasks, performing exceptionally well even in few-shot and zero-shot settings \cite{radford2019language}. However, their potential in time series analysis remains largely unexplored.

Recent studies, including Timer \cite{liutimer}, Time-LLM \cite{jintime}, and LLM4TS \cite{chang2024llm4ts}, have attempted to introduce LLMs into time series tasks. However, these approaches primarily focus on feature alignment between time series data and language models while overlooking semantic task descriptions. Consequently, they fail to fully leverage LLMs' capabilities and lack adaptability to multi-task scenarios.

To address these challenges, we propose LTM (Language Time-series Model), a novel multi-task framework that integrates time series models, LLMs, and knowledge graphs to handle time series tasks, 
including forecasting, imputation, and anomaly detection. LTM first encodes time series data into multiple temporal patches and enhances user prompts using knowledge graphs and descriptive insights derived from the time series itself (e.g., trends and extreme values).

To effectively integrate time series data with textual descriptions, we introduce a feature fusion approach. First, the time series data is segmented and encoded. Simultaneously, textual features are embedded via the LLM's embedding layer and refined using attention pooling. The pooled text features are then fused into each temporal segment, with residual connections preserving the integrity of the time series. Finally, the fused representations are combined with semantic prefixes and fed into the backbone of the LLM, with only the outputs of the time segments used for downstream tasks. To optimize performance across different tasks, we employ task-specific loss functions to guide model training.

Comprehensive evaluations demonstrate that LTM outperforms state-of-the-art time series models. By enhancing time series data through LLMs, LTM provides a robust solution for multi-task scenarios. The contributions of our paper are summarized in four folds:
 \begin{enumerate}
     \item We propose a multi-task time series framework (LTM) that supports multi-modal input and is applicable to a wide range of time series analysis tasks. Compared to SOTA methods, LTM achieves superior performance with fewer parameters and better adaptability. Extensive experiments on various datasets demonstrate the effectiveness of LTM, with a 4\% reduction in average forecasting error in time series forecasting tasks.
     \item We propose a new feature fusion method, the Fusion-Aware Temporal Module (FATM), which enables deep integration of semantic prompts with time series data. FATM effectively captures semantic information while preserving the integrity of time series data. With a simple yet efficient design, it enhances final forecasting performance.
     \item We develop a new prompt enhancement module, Knowledge-Driven Temporal Prompt (KDTP). By incorporating knowledge graphs, KDTP generates high-quality temporal prompts, further enriching semantic understanding. This module is plug-and-play and can be seamlessly integrated into other multi-modal time series frameworks in the future.
 \end{enumerate}

\section{Related Work}

\subsection{Traditional Time Series Analysis Methods}
Traditional time series methods, such as  ARIMA \cite{box2015time}, exponential smoothing \cite{billah2006exponential}, and state-space models  \cite{kalman1960new}, perform well under specific conditions but require stringent assumptions like stationarity or extensive preprocessing. These limitations hinder their ability to capture complex nonlinear relationships and long-term dependencies, reducing their effectiveness for intricate tasks. 
In contrast, deep learning has revolutionized time series analysis by leveraging sequence models to address these challenges. RNN \cite{rumelhart1986learning}, LSTM\cite{hochreiter1997long}, and GRU\cite{cho2014learning}effectively model long-term dependencies and have achieved significant success across diverse tasks. More recently, Transformer-based architectures, including Temporal Fusion Transformer   \cite{lim2021temporal},SDformer\cite{zhou2024sdformer} ETSformer\cite{woo2022etsformer}, Informer\cite{zhou2021informer}, Autoformer\cite{wu2021autoformer}, and TimesNet\cite{wu2022timesnet} have demonstrated superior predictive performance. However, these models often depend on large-scale labeled datasets and face challenges in interpreting complex semantics, particularly in low-data or few-shot scenarios. 
\subsection{Large Time Series Models }
LLM, such as GPT\cite{radford2019language} and LLaMA\cite{tan2024language} have significantly advanced natural language processing, opening new avenues for time series analysis. Emerging research on LTSM primarily focuses on leveraging LLMs for time series tasks, such as feature alignment (e.g., LLM4TS  \cite{chang2024llm4ts}, TIME-FFM\cite{liu2024time} and AutoTimes\cite{liu2024autotimes}), and pretraining on large-scale time series data for tasks like prediction, completion, and anomaly detection (e.g., Timer  \cite{liutimer}, LPTM\cite{kamarthi2023large}and TTM \cite{ekambaram2024tiny} ). Despite progress, as Tan et al.\cite{tan2024language}point out, these approaches often underutilize LLM capabilities, focusing mainly on feature alignment and neglecting semantic information, which limits model generalizability and applicability. 

\subsection{Knowledge Graphs in Time Series}
Knowledge graphs (KGs), as structured representations, capture complex entity relationships and aid contextual understanding of time series data. Recent studies \cite{cai2022temporal,gravina2024temporal,deng2021graph} leverage KGs for prediction and anomaly detection by integrating entity and relationship data. However, most approaches treat KGs as auxiliary information, lacking deep integration of temporal and textual features. Manual KG construction also imposes high costs and scalability challenges, limiting their utility in multimodal temporal tasks.

\section{Method}
Tan et al.\cite{tan2024language} found that existing LLM-based methods primarily focus on time series feature alignment, with limited performance gains attributable to LLM capabilities. Moreover, due to the scarcity of multimodal time series data, current pretraining approaches are confined to the time series modality, restricting their generalization.

\begin{figure*}[t]
    \centering
    \includegraphics[width=\textwidth]{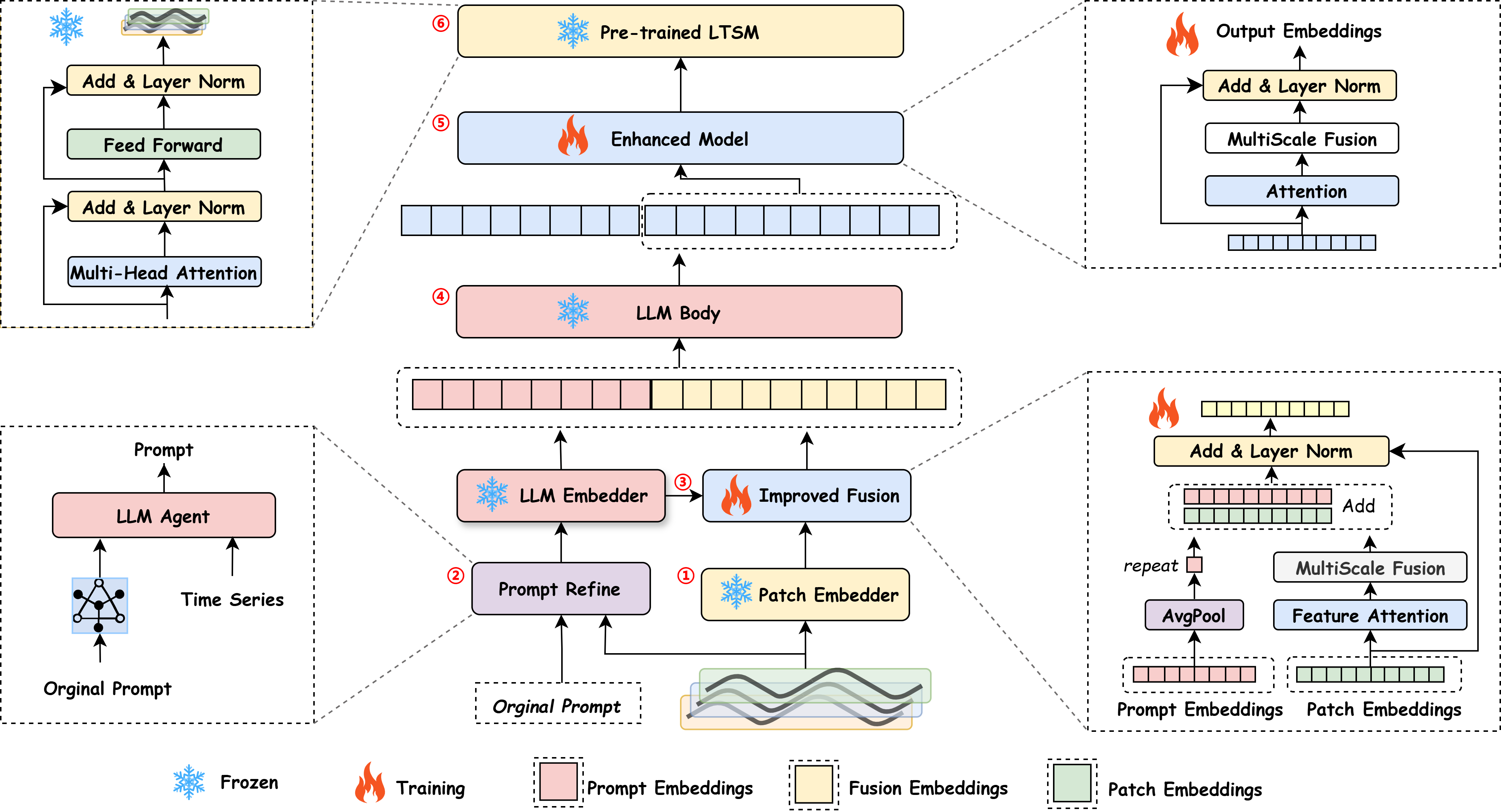}
    \caption{
    The framework of LTM. \textbf{(1)Input Embedding Module:} Converts time series into discrete tokens; \textbf{(2)Knowledge-Driven Temporal Prompt (KDTP):}Generates semantic-rich instructions using task-specific documents; \textbf{(3)Fusion-Aware Temporal Module (FATM):} Fuses temporal and textual features for better task alignment; (4)\textbf{Frozen Pre-trained LLM Module:} Utilizes a frozen LLM backbone for efficient processing; (5)\textbf{Feature Enhancement Module:} Refines fused features to improve downstream task performance; (6)\textbf{Pre-trained LTSM:} Augments temporal modeling without additional training overhead.}
    \label{fig:enter-label1}
\end{figure*}

To address these limitations, we propose an efficient modality fusion framework to fully exploit the capabilities of LLMs. Additionally, we integrate knowledge graphs to reduce semantic noise and enhance time series semantic representation. Our framework, shown in Figure~\ref{fig:enter-label1}, leverages large foundational language models (e.g., LLaMA \cite{touvron2023llama}, BERT \cite{kenton2019bert}, and GPT-2\cite{radford2019language} as boosters for time series tasks. The core of our framework includes a Knowledge-Driven Temporal Prompt (KDTP)
 for extracting task-relevant semantics and a Fusion-Aware Temporal Module (FATM) to optimize instruction guidance.
   
Our framework processes user-provided instructions by generating discrete time series tokens and augmenting initial instructions with knowledge graph-enhanced semantics. During training stage, only lightweight modules, such as feature fusion and output projection layers, are updated, while both the LLM and LTSM backbones remain frozen. This approach ensures efficiency, requiring minimal parameter updates compared to vision-language or multimodal models that often rely on paired cross-modal data for fine-tuning.
Additionally, our framework is resource-efficient, which avoids the need to train large domain-specific models from scratch or perform extensive fine-tuning. Techniques like quantization and compression
 \cite{zhu2024survey} can further reduce memory usage, enabling streamlined deployment without compromising performance. The resulting model demonstrates strong versatility across diverse time series tasks with minimal resource constraints.

\subsection{Input Embedding Module}
First, the input ${{{x}}_{{enc}}}$ is normalized,  by means of z-score. Next, the sequence is divided into consecutive overlapping or non-overlapping segments.This segmentation serves three purposes: (1) Aggregating temporal information from multiple time steps within each segment preserves local temporal patterns; (2) Each segment fuses temporal and semantic information more effectively; (3) As a tokenization step, it compacts the input sequence, reducing computational complexity. 
 \begin{figure*}[t]
    \centering
    \includegraphics[width=\textwidth]{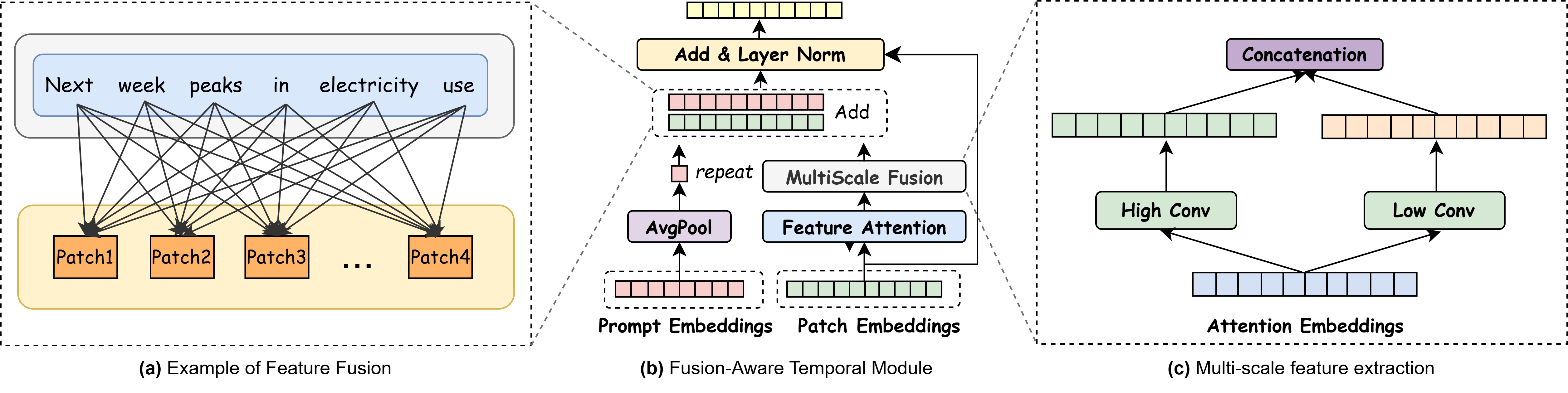}
    \caption{Fusion-Aware Temporal Module.}
    \label{fig:enter-label}
\end{figure*}
\subsection{Fusion-Aware Temporal Module (FATM)}
This module deeply integrates natural language and time series modalities to enhance the model's ability to learn complex temporal patterns. By leveraging multimodal complementarity, the model achieves improved representation, generalization, interpretability, and robustness.
A common approach is to guide LLM to process time series via prompting, suitable for simple sequences with short time steps. For instance, LLMTime
 \cite{gruver2024large} processes time series data as text, utilizing LLMs’ reasoning abilities, while methods like Auto-TTE \cite{chung2023text} convert time series into discrete representations for feature extraction.

However, these methods face limitations with complex multimodal data. To address this, we propose a multi-scale fine-grained feature fusion strategy. Figure~\ref{fig:enter-label}(a) illustrates the incorporation of natural language prompts (e.g., "Next week may see a peak in electricity consumption") into local time series data. This integration enhances the model's understanding of specific scenarios and improves prediction accuracy.
First, natural language prompt embeddings
  ${{P}}$, undergo mean pooling to generate a global prompt representation, as shown in Figure ~\ref{fig:enter-label}(b):
\begin{equation}
    \mathbf{P}_{\text{mean}} = \frac{1}{T_p} \sum_{t=1}^{T_p} \text{softmax}(\mathbf{P}\mathbf{W} + \mathbf{b})_t \cdot \mathbf{P}_t
\end{equation}
where ${{{T}_{{p}}}}$ is the prompt length. This reduces redundancy by summarizing global information. The pooled representation is then repeated to align with the time series length  ${{n}}$ :
\begin{equation}
    \mathbf{P}^r = \text{Repeat}(\mathbf{P}_{\text{mean}}, n)
\end{equation}

This alignment method makes the prompt embedding match each time step of the time series in the time dimension, ensuring that the natural language prompt information can be effectively fused with the time series features.

Next, multi-scale convolutional operations extract time series features at different temporal scales \cite{cui2016multi}. This structure enables the model to observe time series representations across various scales: smaller scales capture local patterns (e.g., daily variations), while larger scales highlight global trends (e.g., weekly or monthly changes). As illustrated in Figure~\ref{fig:enter-label}(c), we employ convolution kernels of different sizes to extract features from diverse temporal windows, enabling comprehensive feature representation.

The fused features integrate representations from both language prompts and time series through a linear transformation, achieving an efficient combination of the two. A learnable weight dynamically balances their contributions, allowing the model to effectively capture complex temporal patterns while maintaining an understanding of natural language. This approach significantly enhances the model's ability to address intricate time series tasks.
\subsection{Knowledge-Driven Temporal Prompt (KDTP)}

\begin{figure*}[t]
    \centering
    \includegraphics[width=\textwidth]{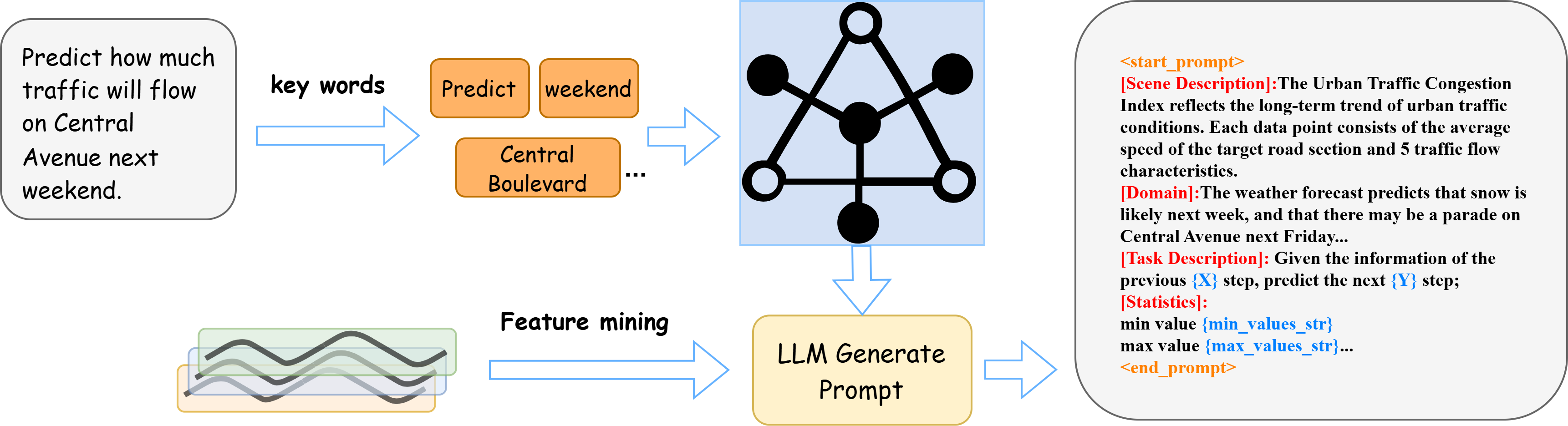}
    \caption{Instruction-Enhanced Framework.}
    \label{fig:enter-label2}
\end{figure*}

In this module, a graph structure based on GraphRAG \cite{edge2024local} combines the user's initial instruction with documents retrieved from an external knowledge database to generate enriched instructions. The RAG framework includes two components: retrieval, which identifies relevant documents using algorithms like cosine similarity and Euclidean distance, and generation, which integrates retrieved information with the query to refine instructions. Formally, the framework is defined as:
\begin{equation}
M = (G, R = (I, S))
\end{equation}
The data indexer \(I\) processes the external database \(D\) and converts it into an indexed form \(\widehat{D}\) for efficient retrieval, that is, \(\widehat{D} = I(D)\). The generation module \(G\) generates the final output based on the input query \(P\) and the results of retrieved documents \(D\). The process from query to generation can be defined as:
\begin{equation}
M(P; D) = G(P, S(P; \widehat{D}))
\end{equation}
The knowledge graph \(M\) integrates task relevant documents from external sources such as social media, weather forecasts, and user-provided files. 
As shown in Figure~\ref{fig:enter-label2}, keywords from the user's query (e.g., predict, weekend, Central Avenue) are used for content-based retrieval. The retrieved documents, relevant to the time series context, are enriched with features such as trends, periodicity, and outliers. The generation module \(G\) then utilizes this information to produce task specific guidance for time series analysis.

\subsection{Feature Enhancement Module}
After processing the enhanced prompt and time series data with the frozen LLM, the prefix prompt is discarded. The outputs are then mapped and enhanced through dimensionality reduction before being fed into the time series decoder to generate the final predictions, as illustrated in Figure~\ref{fig:enter-label}.
\subsection{Multi-Task Learning}
LTM supports diverse time-series tasks, including forecasting, imputation, and anomaly detection.

Forecasting: The forecasting  task is framed as a next-token prediction problem. During fine-tuning, an autoregressive objective is used, where the time series length  ${{L}}$ is divided into ${{T}}$ tokens. The model predicts the next token sequence  $\{\hat{y}_1, \hat{y}_2, \dots, \hat{y}_T\}$, and the loss is computed using MSE:
\begin{equation}
    \mathcal{L}_{forecasting} = \frac{1}{T} \sum_{t=1}^{T} (\hat{y}_t - y_t)^2
\end{equation}

During inference, predictions are iteratively generated by appending each predicted token to the input sequence until the desired length   ${{L + H}}$is reached. 

Imputation: Missing values are treated as masked tokens. The input time series ${{X}}$is combined with a mask matrix  ${{Mask}\in\{0,1\}}$ to form  ${{{X}}_{{inp}}={X}\odot{Mask}}$. The model generates predictions for masked positions, minimizing reconstruction error: 
\begin{equation}
    \mathcal{L}_{imputation} = \frac{1}{\| \text{Mask}_m \|_0} \sum \text{Mask}_m \odot (\mathbf{X} - \hat{\mathbf{X}})^2
\end{equation}
where ${{{Mask}}_{{m}}=1 - {Mask}}$  and ${\left\|{{Mask}}_{{m}}\right\|_{{0}}}$ is the count of masked elements.

Anomaly Detection: Anomalies are detected by comparing predicted future segments to actual observations. Let the observed segment \(X_{obs} = \{x_{t-k}, \dots, x_{t-1}\}\)  predict the future segment  $\hat{\mathbf{X}}_{\text{pred}} = \{\hat{x}_t, \dots, \hat{x}_{t+m}\}$ using the model  ${{{f}}_{\theta}}$ , which learns the conditional distribution ${{P(}{{X}}_{{pred}}|{{X}}_{{obs}}{)}}$ :
\begin{equation}
    \hat{\mathbf{X}}_{\text{pred}} = f_{\theta}(\mathbf{X}_{\text{obs}})
\end{equation}
The reconstruction error between the predicted and true values of the masked segment is calculated as: 
\begin{equation}
    \mathcal{L}_{anomaly} = \frac{1}{T} \sum_{t=1}^T (\hat{x}_t - x_t)^2
\end{equation}
Anomaly scores are defined as \( S_t = d(\hat{\mathbf{X}}_{\text{pred}}, \mathbf{X}_{\text{true}}) \), where \( d \) is the MSE. If \( S_t > \tau \) (threshold), \( x_t \) is flagged as an anomaly:

\begin{equation}
    \text{Anomaly}(t) =
    \begin{cases} 
        1, & \text{if } S_t > \tau, \\
        0, & \text{otherwise}.
    \end{cases}
\end{equation}
This approach formulates anomaly detection as a next-token prediction task, making it well-suited for real-time monitoring of time series data.
To enhance the model's understanding of time series data, we incorporate a penalty loss based on cosine similarity between prompts and time-series patches during fine-tuning: 
\begin{equation}
\mathcal{L}_{total}=\mathcal{L}_{reg}+\lambda\cdot\left(1-\frac{1}{n}\sum_{i = 1}^{n}CosSim(P_{i},F_{i})\right)   
\end{equation}
where \(\lambda\) is a hyperparameter with a range of $0\leqslant\lambda\leqslant1$. This range allows for flexible adjustment of the influence of the penalty loss term. A smaller \(\lambda\) gives more weight to the primary loss function \(\mathcal{L}_{reg}\), while a larger \(\lambda\) emphasizes the importance of the cosine similarity based penalty loss. \( L_{reg} \) represents the primary loss function for different tasks:

\begin{equation}
    \mathcal{L}_{reg} =
    \begin{cases} 
        \mathcal{L}_{forecasting}, & \text{forecasting task}, \\
        \mathcal{L}_{imputation}, & \text{imputation task}, \\
        \mathcal{L}_{anomaly}, & \text{anomaly detection task}.
    \end{cases}
\end{equation}

The term \( 1 - \text{CosSim}(P, F) \) encourages the alignment between the Prompt and the fused features, enhancing the model's perception of both prompt guidance and time series characteristics.
\begin{table*}
    \centering
    \begin{tabular}{p{1.5cm}p{2cm}p{2cm}p{2cm}p{2cm}p{2cm}p{2cm}p{2cm}}
        \toprule
        Methods &\textbf{ LTM}& TimeLLM  & TEMPO  & FEDformer& PatchTST& DLinear & TimesNet \\
        \midrule
        Metric & MAE/MSE & MAE/MSE & MAE/MSE & MAE/MSE & MAE/MSE & MAE/MSE & MAE/MSE \\
        \midrule
        ETTh1 & \textbf{0.376/0.406} & 0.403/0.428 & 0.400/0.406 & 0.509/0.502 & 0.570/0.518 & 0.414/0.421 & 0.407/0.423 \\
        ETTh2 & \textbf{0.288/0.342} & 0.324/0.373 & 0.301/0.353 & 0.385/0.426 & 0.379/0.412 & 0.334/0.389 & 0.315/0.362 \\
        ETTm1 & \textbf{0.284/0.343} & 0.328/0.373 & 0.438/0.424 & 0.698/0.553 & 0.733/0.554 & 0.624/0.522 & 0.518/0.470 \\
        ETTm2 & \textbf{0.176/0.256} & 0.250/0.310 & 0.185/0.267 & 0.665/0.634 & 0.273/0.345 & 0.264/0.352 & 0.202/0.290 \\
        Weather & \textbf{0.151/0.197} & 0.162/0.216 & 0.211/0.254 & 0.292/0.346 & 0.247/0.301 & 0.212/0.275 & 0.247/0.295 \\
        ECL & \textbf{0.130/0.221} & 0.158/0.266 & 0.178/0.276 & 0.300/0.399 & 0.489/0.546 & 0.195/0.292 & 0.293/0.369 \\
        Traffic & \textbf{0.371/0.277} & 0.440/0.329 & 0.476/0.343 & 0.835/0.564 & 1.023/0.641 & 0.609/0.424 & 0.585/0.401 \\
        Avg & \textbf{0.254/0.292} & 0.295/0.328 & 0.313/0.332 & 0.526/0.489 & 0.531/0.474 & 0.379/0.382 & 0.367/0.373 \\
        \bottomrule
    \end{tabular}
    \caption{Long-term forecasting results on time series benchmark datasets. The forecasting length is 96.}
    \label{tab:long-term-results}
\end{table*}
\section{Experiment}
\subsection{Dataset and Experimental Setup}
\paragraph{Datasets}
For time series forecasting, we used widely recognized benchmark datasets: Weather, Traffic, Electricity, and four Electric Transformer Temperature (ETT) datasets as detailed in \cite{wu2022timesnet}. For anomaly detection, we adopted the UCR Anomaly Archive dataset  \cite{wu2021current}, containing 250 tasks. For imputation experiments, we used the PEMS series dataset  \cite{liu2023itransformer}. Appendix A provides detailed statistics for these datasets. 
\paragraph{Baseline Models}
We adopted GPT-2\cite{radford2019language} as the backbone architecture for LTM, illustrated in Figure ~\ref{fig:enter-label1}. To comprehensively evaluate performance, we compared our model with state-of-the-art time series models, referencing reported results from  \cite{cao2023tempo}. The baseline models included:
(1) LLM-based methods: TIME-LLM \cite{jintime} and TEMPO \cite{cao2023tempo}.
(2) Transformer-based models: PatchTST \cite{nie2022time} and FEDformer \cite{zhou2022fedformer}.
(3) Linear model: DLinear \cite{zeng2023transformers}.
(4) Generalized 2D models: TimesNet \cite{wu2022timesnet}.
To ensure fairness under resource constraints, both TIME-LLM and our backbone employed GPT-2 as the underlying architecture. 

\subsection{Long-Term Forecasting}

\paragraph{Setups}
We evaluated long-term forecasting performance on benchmark datasets following \cite{wu2022timesnet}. Details about datasets and implementation can be found in the appendix.
\paragraph{Results}
Our results, summarized in table~\ref{tab:long-term-results}, show that LTM outperforms all baseline models in most cases, demonstrating notable advantages. In particular, LTM achieved average performance improvements of 3.8\% and 4.9\% over Time-LLM  \cite{jintime} and TEMPO, respectively. Time-LLM, a recent method leveraging LLM reprogramming, has shown strong results in long-term forecasting, 
yet LTM still surpasses it significantly. Additionally, compared to traditional models like DLinear, LTM delivered improvements exceeding 10\%.
\subsection{Few-Shot Forecasting}

\paragraph{Setups}
To evaluate LTM's performance under few-shot conditions, we adopted the setup from  \cite{zhou2023one}, using only 5\% of the training time steps. 

\paragraph{Results}
Table~\ref{tab:few-shot-results} shows the few-shot forecasting  results using 5\% of the training data. LTM significantly outperforms all baseline methods, with particularly strong results on the ECL and Traffic datasets. This highlights LTM's ability to activate and effectively utilize the latent knowledge of large models. Under the few-shot setting, LTM’s average MAE and MSE decrease by only 0.037 and 0.011, respectively, surpassing Time-LLM and other baseline methods.This demonstrates the adaptability of LTM in scenarios of extreme data scarcity.
\begin{table}[t]
    \centering
    \begin{tabular}{p{1.3cm}p{1.5cm}p{1.5cm}p{1.5cm}}
        \toprule
        Datasets & ECL & Traffic & Weather \\
        \midrule
        Metric & MAE/MSE & MAE/MSE & MAE/MSE \\
        \midrule
        \textbf{LTM} & \textbf{0.140/0.237} & \textbf{0.391/0.297} & 0.233/\textbf{0.262} \\
        TimeLLM & 0.179/0.258 & 0.421/0.298 & 0.259/0.309 \\
        TEMPO & 0.190/0.290 & 0.560/0.411 & \textbf{0.217}/0.268 \\
        FEDformer & 0.352/0.425 & 0.901/0.611 & 0.292/0.339 \\
        PatchTST & 0.509/0.571 & 1.051/0.674 & 0.291/0.340 \\
        DLinear & 0.205/0.304 & 0.626/0.446 & 0.227/0.283 \\
        TimesNet & 0.503/0.527 & 0.832/0.558 & 0.286/0.326 \\
        \bottomrule
    \end{tabular}
    \caption{Few-shot prediction results based on 5\% of training data.}
    \label{tab:few-shot-results}
\end{table}
\begin{figure}[h]
    \centering
    \includegraphics[width=0.45\textwidth]{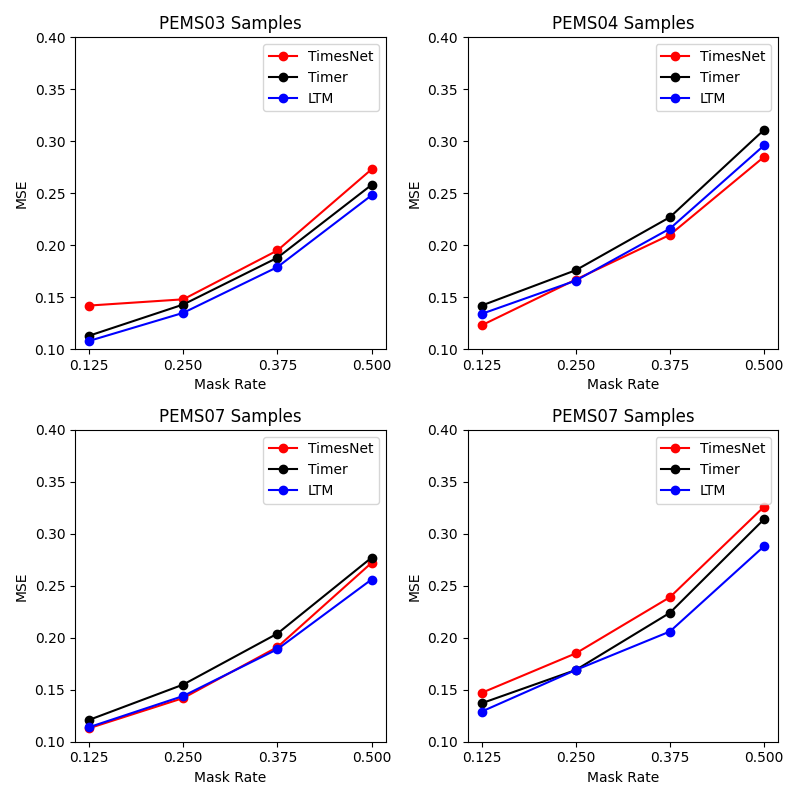}
    \caption{Comparison of the MSE of imputed sequences across different models under varying missing rates on the PEMS datasets.}
    \label{fig:imputation-results}
\end{figure}
\subsection{Time Series Imputation}
\paragraph{Setup}
We followed the setup from Yong Liu et al. \cite{liutimer} and conducted experiments on the PEMS series datasets with missing rates of 12.5\%, 25\%, 37.5\%, and 50\%. The models are 
evaluated by MSE between the imputed and ground-truth sequences. 
\paragraph{Results}
Figure~\ref{fig:imputation-results} illustrates that LTM consistently outperforms baseline methods, including TimesNet and Timer, across all missing rates and datasets. By incorporating scenario description information and multi-scale feature fusion, LTM substantially improves imputation accuracy. LTM significantly enhances imputation accuracy. Under missing rates of 12.5\%, 25\%, 37.5\%, and 50\%, LTM achieves lower MSE than TimesNet and Timer, demonstrating strong adaptability to varying levels of data missingness. 
\subsection{Anomaly Detection}

\begin{figure}[ht]
    \centering
    \includegraphics[width=0.45\textwidth]{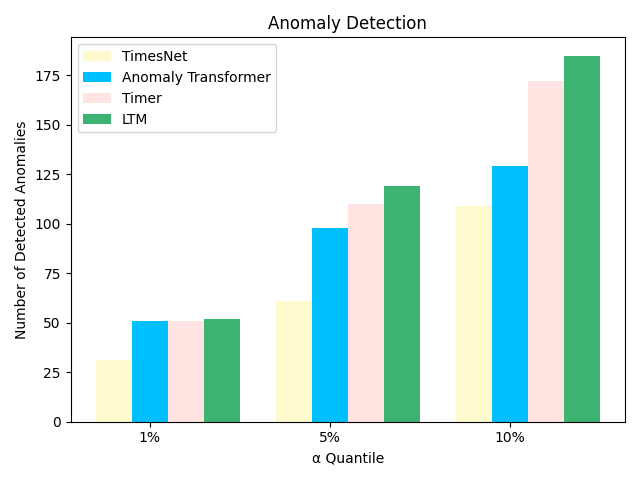}
    \caption{Comparison of the number of anomalies detected by different models at a given confidence quantile on the UCR Anomaly Detection Archive.}
    \label{fig:anomaly-detection}
\end{figure}

\paragraph{Setups}
  We followed the setup in Yong Liu et al.  \cite{liutimer} using the UCR Anomaly Archive dataset \cite{wu2021current}, which comprises 250 tasks. Each task provides a normal time series for training, and the model identifies anomalies in test sequences. After training on the normal sequences, LTM calculates the MSE between predicted and observed values in the test set. Segments with MSE exceeding the \(\alpha\)-quantile threshold are flagged as potential anomalies. 
\paragraph{Results}
As shown in Figure \ref{fig:anomaly-detection}, LTM outperformed well-known models like TimesNet and Anomaly Transformer  \cite{xu2021anomaly}, detecting more anomalies under the same threshold. Furthermore, LTM slightly surpassed the state-of-the-art Timer model, demonstrating the benefits of incorporating temporal descriptive information.

\begin{figure}[h]
    \centering
    \includegraphics[width=0.45\textwidth]{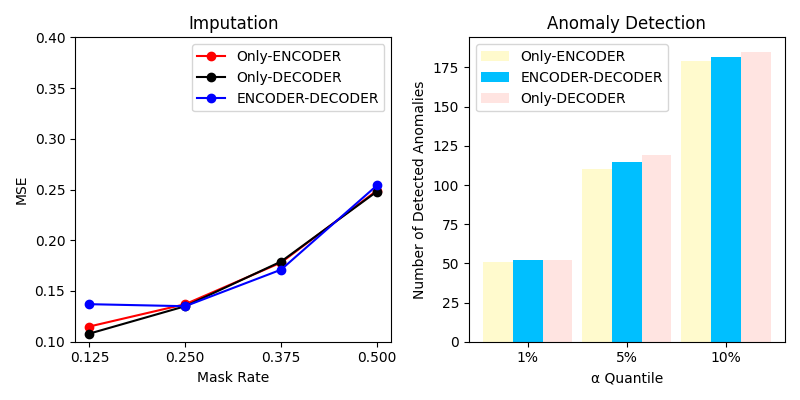}
    \caption{Performance of different transformer models on time series imputation and anomaly detection tasks.}
    \label{fig:model-performance}
\end{figure}

\subsection{Model Analysis}
\paragraph{Variants of Language Models.}
We analyzed the impact of different Transformer architectures on time series tasks by comparing three common structures: encoder-only models (e.g., BERT \cite{kenton2019bert}), decoder-only (e.g., the GPT series\cite{radford2019language}), and encoder-decoder (e.g., BART and T5\cite{lewis2019bart,raffel2020exploring}). The specific models used are detailed in Appendix C. Decoder-only models excel in zero-shot and few-shot learning, encoder-only models are ideal for processing entire input sequences, and encoder-decoder models perform better in complex tasks due to their separation of input and output. 
As shown in Table~\ref{tab:compact-results} and Figure~\ref{fig:model-performance}, the decoder-only GPT-2 model outperformed other architectures across various time series tasks, achieving the lowest MAE and MSE. This demonstrates the compatibility between the decoder-only structure and time series tasks, as its generative modeling aligns naturally with the autoregressive nature of time series data.
\paragraph{Ablation Study}
To evaluate the contribution of each LTM component and the effectiveness of LLM in time series forecasting, we performed ablation studies on the Weather, ETTh1, and ETTm1 datasets. We independently removed Fusion-Aware Temporal Module, Knowledge-Driven Temporal Prompt, and LLM. appendix D visualizes features before and after fusion. \textbf{w/o FATM:} Concatenates time-series description and time-series features without mean-pooling fusion. \textbf{w/o KDTP:} Inputs only time-series features, excluding descriptive enhancements. \textbf{w/o LLM:} Replaces the LLM with a fully connected network.
\begin{table}[H]
    \centering
    \begin{tabular}{lccc} 
        \toprule
        {Methods} & BERT  & GPT2 & T5 \\
        \midrule
        Metric& MAE MSE & MAE MSE & MAE MSE \\
        \midrule
        ETTh1   & 0.381 0.410 & \textbf{0.376 0.406} & 0.385 0.410 \\
        ETTh2   & 0.320 0.370 & \textbf{0.288 0.342} & 0.323 0.371 \\
        ETTm1   & 0.290 0.349 & 0.284 0.343 & \textbf{0.280 0.340} \\
        ETTm2   & \textbf{0.174 0.257} & 0.176 0.256 & 0.180 0.258 \\
        Weather & \textbf{0.146 0.188} & 0.151 0.197 & 0.153 0.198 \\
        ECL     & 0.132 0.229 & \textbf{0.130 0.221} & 0.133 0.225 \\
        Traffic & 0.378 0.281 & \textbf{0.371 0.277} & 0.380 0.289 \\
        Avg     & 0.260 0.298 & \textbf{0.254 0.292} & 0.262 0.299 \\
        \bottomrule
    \end{tabular}
    \caption{Performance comparison of different transformer models across datasets. }
    \label{tab:compact-results}
\end{table}

\begin{table}[t]
    \centering
    \begin{tabular}{p{0.9cm}p{1.5cm}p{1.6cm}p{1.6cm}p{1.5cm}}
        \toprule
        Methods & \hspace{1.5em}LTM& w/o FATM & w/o KDTP & w/o LLM \\
        \midrule
        Metric & MAE/MSE & MAE/MSE & MAE/MSE & MAE/MSE \\
        \midrule
        ETTh1 & \textbf{0.376/0.406} & 0.395/0.425 & 0.406/0.428 & 0.390/0.412 \\
        ETTm1 & \textbf{0.284/0.343} & 0.331/0.376 & 0.335/0.380 & 0.330/0.376 \\
        Weather & \textbf{0.151/0.197} & 0.171/0.213 & 0.179/0.222 & 0.169/0.210 \\
        Avg & \textbf{0.260/0.298} & 0.299/0.338 & 0.307/0.343 & 0.296/0.333 \\
        \bottomrule
    \end{tabular}
    \caption{Results of the ablation study across different LLM modes.}
    \label{tab:ablation-results}
\end{table}

Table~\ref{tab:ablation-results} shows that the complete LTM outperforms all ablated versions, confirming the significance of each module. Removing FATM reduces MAE by 3.9\%, demonstrating the importance of mean-pooling fusion for feature interaction. Excluding KDTP causes a 3\% MAE drop on ETTh1, underlining the value of descriptive features. Although the impact of removing LLM is smaller, it still improves prediction.
\paragraph{Model Efficiency}
To evaluate LTM's efficiency in fine-tuning resource consumption and inference speed, we compared it with TIME-LLM, the second-best model in predictive performance, as shown in Table \ref{tab:efficiency-analysis}. LTM activates LLM for time series analysis with fewer than 11 million trainable parameters, representing approximately 7\% of the total model size. Its inference speed is nearly twice that of TIME-LLM. Removing the LLM further improves inference efficiency, highlighting that LTM's overall efficiency is largely determined by its base language model.

\begin{table}[H]
    \centering
    \sisetup{table-format=3.4} 
    \begin{tabular}{l S S S} 
        \toprule
        Methods & {Param(M)} & {Train.Param(M)} & {Speed(s/iter)}\\
        \midrule
        Time-LLM   & 132.9635 & 51.0509 & 1.3218 \\
        LTM        & 145.8957 & 11.5430 & 0.9342 \\
        w/o LLM    &  11.5430 & 11.5430 & 0.5632 \\
        \bottomrule
    \end{tabular}
    \caption{Efficiency analysis when forecasting on the ETTh1 dataset.}
    \label{tab:efficiency-analysis}
\end{table}

\section{\textbf{Conclusion}}
This paper proposes LTM, a multi-task time series framework integrating LLMs, pre-trained time series models, and knowledge graphs for deep fusion of temporal and semantic features. Specifically, we introduce the Fusion-Aware Temporal Module to integrate semantic prompts with time series representations, preserving temporal integrity while enhancing semantic expressiveness. Additionally, we propose the Knowledge-Driven Temporal Prompt module, leveraging knowledge graphs to generate high-quality time-aware prompts, enriching contextual understanding in time series tasks. These innovations empower LTM to handle forecasting, imputation, and anomaly detection with enhanced adaptability and efficiency. Extensive experiments on multiple benchmark datasets demonstrate that LTM outperforms the SOTA methods while maintaining high computational efficiency. Future research will explore LTM's scalability to large scale real world applications and further optimize its real-time processing capabilities.

\bibliographystyle{named}
\bibliography{ijcai25}

\end{document}